\documentclass[letterpaper]{article} %
\usepackage{aaai23}  %
\usepackage{times}  %
\usepackage{helvet}  %
\usepackage{courier}  %
\usepackage[hyphens]{url}  %
\usepackage{graphicx} %
\usepackage{natbib}  %
\usepackage{caption} %
\usepackage{algorithm}
\usepackage{algorithmic}
\usepackage{amsmath}
\usepackage{amssymb,enumitem}
\usepackage{newfloat}
\usepackage{listings}
\DeclareCaptionStyle{ruled}{labelfont=normalfont,labelsep=colon,strut=off} %
\floatstyle{ruled}
\newfloat{listing}{tb}{lst}{}
\floatname{listing}{Listing}
\title{Ambient Adventures: Teaching ChatGPT on Developing Complex Stories}
\author{
    Zexin Chen\equalcontrib,
    Eric Zhou\equalcontrib,
    Kenneth Eaton,
    Xiangyu Peng,
    Mark Riedl
}
\affiliations{
    Georgia Institute of Technology, Atlanta, GA, 30332, USA
}

\usepackage{bibentry}

\usepackage{tikz}
\usepackage{enumitem}

\definecolor{brightcerulean}{rgb}{0.11, 0.67, 0.84}

\usepackage{graphicx,calc}
\newlength\myheight
\newlength\mydepth
\settototalheight\myheight{Xygp}
\usepackage{amssymb}

\usepackage{amsmath}
\usepackage{paralist, tabularx}
\usepackage[normalem]{ulem}
\usepackage{latexsym}
\usepackage{multirow}
\usepackage{booktabs}
\usepackage[normalem]{ulem}

\usepackage{enumitem}

\begin{document}

\maketitle

\begin{abstract}

Imaginative play is an area of creativity that could allow robots to engage with the world around them in a much more personified way. 
Imaginary play can be seen as taking real objects and locations and using them as imaginary objects and locations in virtual scenarios.
We adopted the story generation capability of large language models (LLMs) to obtain the stories used for imaginary play with human-written prompts. 
Those generated stories will be simplified and mapped into action sequences that can guide the agent in imaginary play. 
To evaluate whether the agent can successfully finish the imaginary play, we also designed a text adventure game to simulate a house as the playground for the agent to interact.

\end{abstract}

\section{Introduction}

In recent years, the domain of agents has experienced extraordinary progress, driving the creation of intelligent machines that connect the realms of science fiction and reality.
As researchers, engineers, and innovators collaborate, the evolution of agents keeps pushing the limits of technology.
However, how do we ensure that agents have a persistent, yet non-intrusive presence in the household? Considering kids: they are never idle --- they find ways to occupy their time through play and if that play is imaginative play, then the entire home becomes a playground. 
We propose to develop the computational capability for agents to engage in imaginative play and link that play to navigation through the home. This will increase the presence of the agent in the home without directly demanding attention from people, but also using curiosity to invite engagement.

Imaginative play is an exemplar of everyday human creativity in which real-world, mundane objects and locations act as substitutes for imaginary objects and locations as part of a pretend scenario\cite{zook2011formally}. A terrarium can be a garden for growing magic seeds, a kitchen can be a laboratory, or a broom handle can be a light saber. Imaginative play is fundamental to human creativity. Computational systems that can engage in imaginative play can create a sense of presence and persona and provide opportunities for improvisational interactions.

In this paper, we are focusing on exploring how to guide an agent to execute imaginary play with large language models such as ChatGPT \cite{chatgpt}. 
Text adventure games serve as useful test beds because they have also been demonstrated to transfer to visual and real-world domains~\cite{Wang2022Science,ALFWorld20,peng2022inherently}.

\section{Related Work}

ext games are turn-based games where players read descriptions of the current scene for information and interact with short descriptions of actions \cite{cote2018textworld, Wang2022Science, ALFWorld20}. During the  designing of the text adventure game, we follow the text game structure \cite{peng2023story} to create a house consisting of rooms that contain objects as the realistic mappings of the imaginary play. The difference is that our adventure games don't have NPCs since we want to focus on whether the story of imaginary play can guide the agent instead of elevating the difficulty of interaction in the game. 

ChatGPT is an LLM chatbot developed by OpenAI that can be interacted with in multiple ways, including giving prompts to write stories \cite{chatgpt}. Many pre-trained LLMs have had success in story generation; by decomposing textual story data into a series of events, it has been found that these models can generate stories from these events that are more coherent and logical \cite{Martin_2018,peng2022guiding, peng2022inferring}. We aim to adopt this ability to train the model with prompting and generate stories that can guide the agent in imaginary play with ChatGPT.

\begin{figure*}
  \centering
    \includegraphics[width=\textwidth]{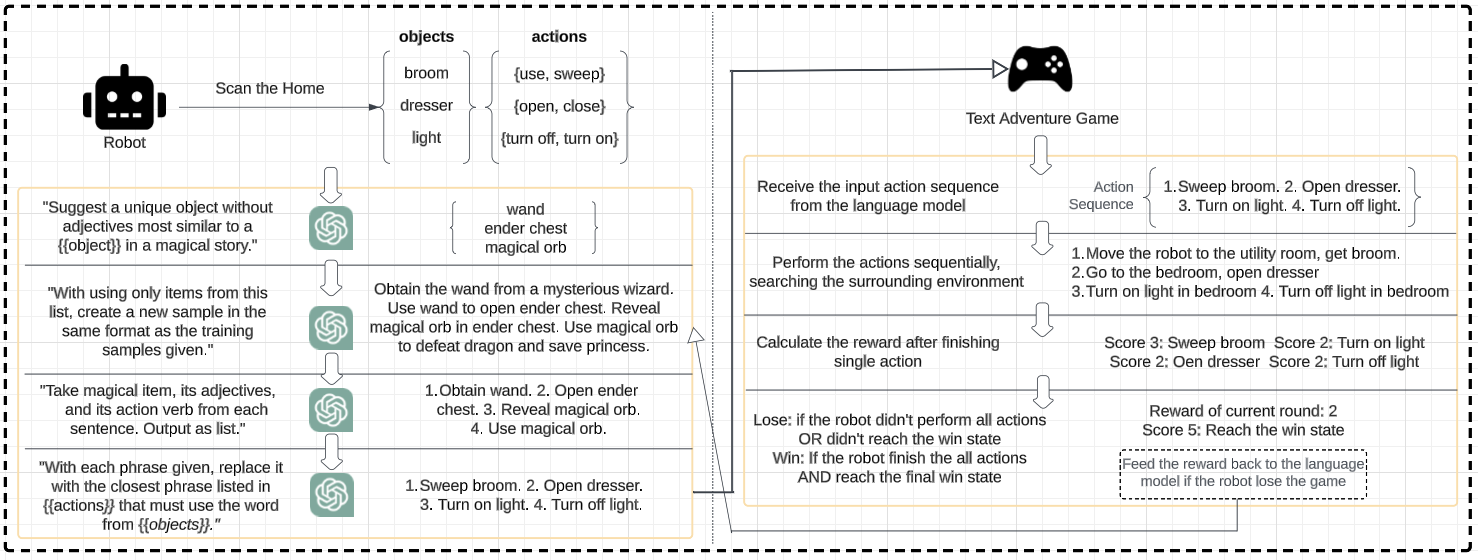}
    \caption{Pipeline Architecture for Text Game and ChatGPT. A sample iteration is demonstrated.}
    \label{fig:pipeline}
\end{figure*}

\section{Story Generation} \label{GPTGeneration}

A natural language story is generated automatically as an exemplar of the behavior the agent is to enact in imaginary play.
Given the topic of the imaginary play, Large Language Models (LLMs) such as ChatGPT and GPT-4 are used to create the imaginary story, and the real-world objects are transformed into similar objects in an imaginary world --- \textit{imaginary objects}, to facilitate the imaginary play.

\subsection{Imaginary Story Generation}
Firstly, the agent scans the layout of each house to obtain the real-world objects and rooms, as well as their respective locations in the house.
In this process, LLMs (ChatGPT) generate imaginary objects which match the setting of the imaginary world (whether that be magical, horror, etc.) that have similar characteristics to the object it was mapped from.
For example, a ``\texttt{broom}" can be transformed into a ``\texttt{wand}" in the imaginary setting because they have similar shapes, are both made of wood and can be held. 
See Fig~\ref{fig:pipeline}, where the first prompt by ChatGPT is used to map each original object to an imaginary one.

With each real-world object, we also obtain the admissible actions for each one – $A_o$ – an admissible action refers to one that can be performed with that object. 
Let there be a set of real-world objects, such as $\langle$``\texttt{broom}", ``\texttt{dresser}", ``\texttt{mug}"$\rangle$. The set $O_o$ denotes the set of real-world objects.
For example, a ``\texttt{broom}" in the house may have the set of admissible actions $\langle$``\texttt{sweep}", ``\texttt{pick up}"$\rangle$. 
With this, LLMs are prompted to find the closest, most similar imaginary object for each item in $O_o$, based on the setting of the story.
By this, we map all of our objects from the original $O_o$ to the set of imaginary objects --- $O_n$.
Within any story, a topic, such as "saving a princess", is required for LLMs (ChatGPT) to aim to complete. 
A wide variety of topics were given based on the setting. For example, a magical setting could entail saving a princess, or a horror story would entail finding a key to escape.
\begin{table}[h]
    \caption{Magical Story Example}
    \label{tab:ex_actions_chatgpt}
    \begin{tabular}{p{0.9\linewidth}}
    \toprule
    \textbf{Topic: Magical World - Saving a Princess}\\
    \midrule
    \textbf{Imaginary Story (First Iteration):} Whisperweaver discovers hidden passage. Uncover ancient chest in hidden passage. Open chest to reveal enchanted staff. Also find Crescent Mirror in chest. Wield enchanted staff for enhanced spellcasting. Use Crescent Mirror for scrying and divination. Harness the power of the enchanted staff and mirror to defeat evil forces and save princess.\\
    \midrule
    \textbf{Simplified Story:} 1. Discovers Whisperweaver 2. Uncover Ancient Chest 3. Reveal Enchanted Staff 4. Find Crescent Mirror 5. Wield Enchanted Staff 6. Use Crescent Mirror 7. Harness Enchanted Staff.\\
    \midrule
    \textbf{Real-World Translation:} 1. Wear clothes 2. Open nightstand 3. Use broom. 4. Open dresser 5. Use broom. 6. Open dresser 7. Use broom.\\
    \midrule
    \textbf{Imaginary Story (Second Iteration):} Whisperweaver discovers hidden passage. Uncover ancient chest in hidden passage. Open chest to reveal enchanted staff. Also find Crescent Mirror in chest. Wield enchanted staff for enhanced spellcasting. Use Crescent Mirror for scrying and divination. Discover recipe for elixir with Crescent Mirror. Brew elixir in the cauldron. Use enchanted staff to activate the elixir. Use transformed abilities from elixir to defeat the evil threat.\\
    \bottomrule
\end{tabular}
\end{table}

To ensure that the generated story by ChatGPT can be simplified into phrases later, several training samples were given to it to ensure that it would be easy to do so. These training samples are short, 5-7 sentence imaginary stories that are concise and contain several random imaginary objects and a topic (such as "defeating the dragon").
Refer to Fig~\ref{fig:pipeline} and the second ChatGPT prompt which uses these training samples and a list of the imaginary objects to generate new samples to use. 
A common limitation of ChatGPT's story generation is that it will simply create a story where the agent immediately obtains every imaginary item in one sentence and follows by completing the topic.
While this indeed works logically, it is far from interesting. 
With the training samples given, ChatGPT is prompted to at most add one new item in each sentence. 
For example, if ChatGPT is given 5 magical items and a topic to save a princess, it will initially get all 5 items in the first sentence and then save the princess in the second. 
With this restriction, ChatGPT has to find a logical way to use objects to get others and continue until it has enough objects to save the princess.

LLMs may sometimes be prompted to rewrite the story – instead of entirely rewriting the story ChatGPT will be prompted to continue it – refer to Table 1 for the second iteration. A sample iteration would involve the last sentence of the original imaginary story from Table 1 being removed and ChatGPT prompted to generate new sentences starting from this point to reach the intended topic. 
At this point, ChatGPT has successfully generated a story, and now this story needs to be distilled and translated back into admissible actions in the text game.

\subsection{Mapping and Filtering} 

The reason for distilling the story is so that these actions can be given for the text game to easily understand. Each sentence in the imaginary story is distilled into a phrase by taking the one imaginary object in each sentence and the action verb that is associated with it. Refer to Figure~\ref{fig:pipeline} and the third ChatGPT prompt, as well as the \textit{Simplified Story} in Table 1.
If there is more than one object in a sentence, the newly obtained object is chosen. For example, if a sentence is ``\texttt{open chest to reveal staff}", the distilled phrase will be ``\texttt{reveal staff}", not ``\texttt{open chest}". 
Once we have all of the phrases, we now want to map these phrases back into admissible actions that can be performed in the real world.

Remember that $A_o$ refers to the admissible actions for original objects, such as $\langle$``\texttt{sweep}", ``\texttt{use}"$\rangle$ for a broom. ChatGPT can then identify the most similar admissible action in $A_o$ that best matches the action performed in the imaginary world. 
For example, a ``\texttt{broom}" in the real world is mapped to a ``\texttt{wand}" – see Table 1 for an example of an imaginary story.
If the ``\texttt{wand}" is used to cast a spell, ChatGPT would determine which admissible action would be most similar to the action ``\texttt{cast a wand}".
If ``\texttt{sweep}" is chosen, then ``\texttt{cast a wand}" will be mapped into ``\texttt{sweep broom}" – see Table 1 once again and the mapping from \textit{Simplified Story} to \textit{Real-World Translation}.
The agent can then use these mapped admissible actions to interact with the real-world environment. 

\section{Text Adventure Games } \label{sec:methods}

Text Adventure Game is the testbed to show how the agent does imaginary play in the real world. 
Text games show the event happening within the current scene by depicting the existing objects and happened actions in short sentences. Objects taken as entities in the game contain the states and actions used in presetting the interactions with the agent. The story might ignore details when events happen, but the text game can record the hidden state changes with words.  For this reason, we utilize text games as test playgrounds.

We evaluate the performance of an imaginary story by whether the agent can perform all input actions sequentially, which indicates that the model-translated action sequences can function as guidance to the agent in imaginary play. The reward of each round of the game equals the score of the last action. To tell the game result directly, we set the last action in the sequence as the win state. Thus, if the reward equals the score of the preset win state, the agent successfully finishes all actions given in the sequence.

We developed our text game in TextWorld \cite{cote2018textworld}, an open-source, extensible engine that both generates and simulates text games. In the game, we mimic the physical environment by mapping out the house floor plan and including pre-scripted interactions with each object in the room to give guidance under different use cases \cite{ narasimhan2015language}. 

\subsection{Game Design Details}
\label{sec:textgame}
\begin{figure}[h]
\includegraphics[width=\linewidth]{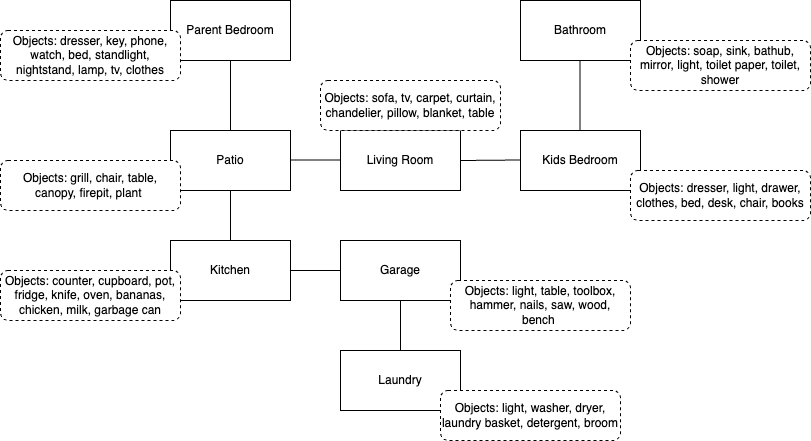}
    \caption{Layout of game ``Housework''.}
    \label{fig:house_map}
\end{figure}

We design a game by inserting a base map that records the location of each room and logic objects related to different rooms (See Fig~\ref{fig:house_map}). 
Each room has its own furniture and appliances, some of which are required, like a light.
Each object $O_i$ has its respective action set $A_i$ which can change the states of itself and the game.
For instance, the object \texttt{Clothes}, $O_{clo}$, has two states: "washed" and "not washed". 
The action \texttt{wash} in an action set $A_{clo}$  can convert the state of $O_{clo}$ from "not washed" to "washed" when the agent successfully finishes the action --- ``\texttt{wash cloth}". To execute this action, the agent starts in the \textit{parentBedroom} to grab the dirty clothes. When it moves to the \textit{laundry} and finishes washing the clothes, the agent gains 2 points for \texttt{wash clothes}. 

For the agent to distinguish similar verbs with similar meanings and reactions, for instance, ``\texttt{wash cloth}" and ``\texttt{clean cloth}" should be taken as the same thing to do. 
In that way, we ensure the agent will take the same reaction every time with synonyms.

\subsection{Reinforcement Learning Agent in Text Game}
\label{sec:game_flow}
The agent has an action sequence that needs to be finished in the text game and obtains rewards for successfully changing the state of the objects or the game.
The game process is as follows: the agent always starts in a fixed room with a given input action sequence. 
The agent will obtain a reward when it finishes the input action by interacting with the surrounding environment. 
From section 4.1, if the agent successfully finishes the action ``\texttt{wash cloth}", it will obtain 2 points as a reward. 
The reward the agent will gain depends on the difficulty of the action. 
We categorize the activities into three levels: stand-alone, interactive, and win, corresponding to 2, 3, and 5 points.

    1. We define \textit{stand-alone actions} as actions that the agent is able to finish without using any other objects.For example, when the agent takes the action ``\texttt{turn on the light}", the \texttt{Light} can be turned on directly after locating the \texttt{Light}. 
    
    2.  The definition of \textit{interactive actions} is the actions that the agent is able to interact with other objects. For example, when the agent takes action ``\texttt{water plant}" and sees the \texttt{Plant}, it cannot directly take action if there is no water in its hands. The agent needs to get the kettle, then check whether the kettle is full of water. If not, it will fill the kettle, then carry it to \texttt{water plant}. 
    
    3.  \textit{Win action} is the last action that the agent needs to finish. When the agent takes the last action in the given action sequence, ``\texttt{clean the oven}'', it comes to the kitchen, locates the \texttt{Oven}, and then \texttt{clean} it. The game ends when the agent successfully goes through the whole action sequence and finishes all actions in it.

\section{Findings}
\label{sec:exp}

\textbf{Story Generation of Large Language Model}:

We used ChatGPT as the LLM to generate stories during our experiment (See Section \ref{GPTGeneration}). 
Most stories required several iterations of revision (Refer to Table 1) until they included the win state in the action sequence. 
Two limitations of the current model are limited prompting formats and difficulty in understanding interactive actions in the text game. 

The first relates to the drawback of the language model is that the generation is uncontrolled. Aside from an initial prompt, generative language models are guided by word co-occurrence, which can lead to repetition, as well as a tendency to focus on descriptive details that do not move a story forward.
To solve the problem, we kept crafting prompts to direct the model to create coherent and executable stories with a clear goal and formed a fixed prompting format. The format limits the adaptivity of the agent to varied types of imaginary play. 
If the setting in imaginary play is modified, the model needs new prompts for the changes. 
           
The other limitation is the difficulty forChatGPT to understand connections between objects in the text game. The generated story cannot associate the objects picked from a previous room with those in the current room if no detailed prompts. That may lead to generating actions not allowed within the text game in the action sequence.
The solution to alleviate such problems is to introduce the missing connections into the prompting and have more iterations of story generation to update the prompt with the generated output.

\paragraph{Game Results with Promptings.}
Results from our sample stories indicate that the agent cannot determine the final win state by itself. To increase the possibility of win, we record the result and feed it back to the language model (ChatGPT). The model knows whether the agent successfully reaches the win state from the previous round's score. If the agent doesn't win, the prompt will tell the model to generate more descriptions of directional information in new stories to guide the agent in the next round.
Although new instructions might not give the expected results every time, we still are able to catch the pattern and re-prompting ChatGPT to better train in zero-shot learning.

\section{Conclusions}

Imaginary play is a creative direction for developing agent
learning abilities. With the help of story generation from LLMs (such as ChatGPT \cite{chatgpt}), we can tell the model to generate imaginary play stories that guide the agent's interactions through prompts. Story generation allows the agent to develop interesting imaginative stories with the objects and topic given, allowing the agent to engage in imaginative play in the real world.

We use text games to model what happens within a given story and the interactions the agent generates with the setting, making the  interaction controllable and explainable. Through mapping imaginative play to real-world scenarios through text games, we figured out how to use rewards to better prompt the model and construct the stories that can guide the agent in imaginary play.

\clearpage
\bibliography{aaai23}

\end{document}